\documentclass[runningheads]{llncs}

\usepackage[year=2026,ID=1]{accv}

\usepackage{accvabbrv} %
\usepackage[numbers,sort&compress]{natbib} %

\usepackage{accvabbrv}

\usepackage{graphicx}
\usepackage{booktabs}

\usepackage[accsupp]{axessibility}  %

\usepackage{makecell}

\usepackage[pagebackref,breaklinks,colorlinks,citecolor=accvblue]{hyperref}

\usepackage{orcidlink}

\usepackage[dvipsnames]{xcolor}
\usepackage{wrapfig}
\usepackage{float}

\newcommand{\video}{\mathcal{V}}          %
\newcommand{\domain}{\Omega}              %
\newcommand{\timeset}{\mathcal{T}}        %
\newcommand{\cut}{\gamma}                 %
\newcommand{\sti}{\mathcal{S}}            %
\newcommand{\cutset}{\mathcal{C}}         %
\newcommand{\pos}{\mathbf{p}}             %
\newcommand{\Fmeas}{F_{1}}

\usepackage{ifthen}
\usepackage{comment}
\newboolean{mythings}
\setboolean{mythings}{true} %

\ifthenelse{
    \boolean{mythings}
}{

}{
    
    \excludecomment{mynote}
    \excludecomment{mywarning}
    \excludecomment{myquestion}
    \excludecomment{mytodo}
    \excludecomment{myexample}

}

\begin{document}

\title{Thinking from Another Side of the Box:\\ Spacetime Images for Computer Vision}

\titlerunning{Spacetime Images for Computer Vision}

\author{Jade Lambert \and
S\'ebastien Pi\'erard\orcidlink{0000-0001-8076-1157} \and
Anthony Cioppa\orcidlink{0000-0002-5314-9015} \and
Marc~Van~Droogenbroeck\orcidlink{0000-0001-6260-6487}}

\authorrunning{Lambert et al.}

\institute{Montefiore Institute, University of Li\`ege, Li\`ege, Belgium\\
\{anthony.cioppa, s.pierard, m.vandroogenbroeck\}@uliege.be}

\maketitle

\begin{abstract}
Computer vision methods for video understanding, such as object detection and tracking, inherit the way videos are captured and displayed: %
as successive frames processed sequentially and 
linked frame-to-frame.
However, a video is an inhomogeneous 2D+time volume of pixels, named voxels, and frames are only one family of cuts through it.
Furthermore, in this frame-based representation, static backgrounds introduce a lot of computational redundancy, and motion can only be understood across several frames. 
In this position paper, we argue that slicing is a design variable and study another family of cuts: \emph{spacetime images} (STIs), obtained by sampling any other set of pixels than frames; for instance, a line of pixels followed over time. %
Such images are familiar in narrow forms, from line-scan inspection and kymographs to spatiotemporal slices for motion analysis, but they have received little attention as a general input representation for computer vision and are often limited to $(x,t)$ or $(y,t)$ cuts.
Specifically, we define new cuts leading to STIs in their general form, explain how to read their content, and show that a family of cuts tiles the video so that existing dense annotations transfer at no cost. We also detail how motion appears in an STI, spatially and in the frequency domain, and discuss applications, such as pedestrians walking, detecting deposited objects, or analyzing dynamic scenes.
We then present two pilot experiments to illustrate the representation's use. 
The first trains a segmentation network entirely in the spacetime domain, on transferred annotations, and performs background subtraction on CDnet~2014 close to supervised frame-based methods, without any explicit temporal modeling.
The second recovers the pan speed of a camera as a moving cut, identified by a simple criterion on the frequency spectrum of the resulting image, which turns a temporal estimation problem into the choice of a cut.
Our position is not that STIs should replace frames, but that the slicing of the video volume is a design variable worth studying in its own right, with advantages and limitations that we discuss.
\end{abstract}

\section{Introduction}

\begin{figure}[htbp]
    \centering
    \includegraphics[width=0.9\textwidth]{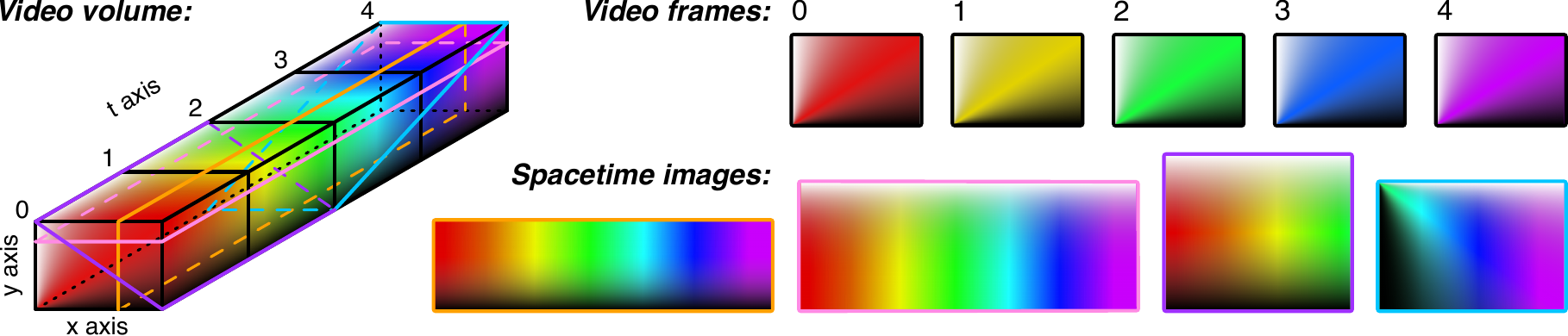}
    \caption{\textbf{Illustration of Spacetime Images (STIs).} \textbf{[Left]} A video is represented as a volume of voxels in the inhomogeneous $(x, y, t)$ coordinate space. \textbf{[Top right]} Conventional video frames correspond to successive $(x, y)$ cuts, read in sequence; here, each frame is a diagonal value gradient from white to a saturated hue to black, with the hue shifting over frames (time). \textbf{[Bottom right]} Spacetime images are obtained from any other cuts. The pink cut, parallel to $x$, yields an $(x, t)$ and the orange cut, parallel to $y$, yields a $(y, t)$ STI, both showing the hue shift. The purple cut is an oblique one in the $(x,y)$ plane but still evolves along $t$, whereas the cyan cut evolves along a direction with a spatial component and therefore samples a different location at every time. %
    }
    \label{fig:XYT cube rainbow}
\end{figure}

Videos are usually represented as temporal sequences of 2D images, but they can equivalently be viewed as three-dimensional volumes in the spatio-temporal coordinates $(x, y, t)$, as illustrated in \cref{fig:XYT cube rainbow}. 
The most intuitive way for humans to visualize and analyze such a volume is to look at it from the front, examining its $xy$ cuts one after the other along the temporal axis. 
These cuts are precisely the video frames, the units in which videos are captured by cameras and displayed on screens.
Consequently, this frame-centric view has heavily shaped the design of computer vision methods across many tasks.
Multi-object tracking, for example, commonly follows the tracking-by-detection paradigm: objects are first detected in each frame, and these detections are then linked across frames.%

However, even though frame-based computer vision methods are effective and constantly improving, they can be computationally expensive and redundant, since each complete image is processed at every time step, even when a static background dominates and only a few pixels move. %
Moreover, motion is not contained in a single frame: frame-based pipelines must reconstruct it from several frames, with optical flow~\cite{Dosovitskiy2015FlowNet}, temporal models~\cite{Tran2015Learning, Liu2022Video} or background models~\cite{Barnich2011ViBe,StCharles2015SuBSENSE}. 
Event cameras~\cite{Gallego2022EventBased} reduce the redundancy at the sensor by reporting only the pixels that changed, but they produce a sparse stream of binary polarity rather than complete images, so that recovering the appearance needed for deeper semantic understanding requires additional reconstruction steps~\cite{Rebecq2019EventsToVideo}.%

In this position paper, we investigate \emph{SpaceTime Images} (STIs) as an alternative representation on which to develop and run computer vision methods.
Unlike conventional video frames, STIs are obtained from other cuts in the video volume: they represent the temporal evolution of a selected set of voxels, for instance a line of pixels, as a single image.
In other words, frames are what one sees when looking at the front of the video volume while STIs are what one sees when looking at it from another side, \textit{i.e.,} from any other cut, as illustrated in \cref{fig:XYT cube rainbow}.\footnote{Throughout this paper, \emph{image} denotes any two-dimensional cut of the video volume, and \emph{frame} only a conventional video frame.}
Specifically, STIs map temporal evolution onto a spatial dimension, so that the dynamics of a whole video are read in a single image.
In such an image, a static background degenerates into parallel streaks, camera motion is written into their deformation, and moving objects stand out as imprints that break them.
Such images are familiar in narrow forms, from line-scan inspection~\cite{Zographos1998Line} and kymographs~\cite{Smal2010Microtubule} to spatio-temporal slices for motion analysis~\cite{Ngo2003Motion}, but they have received little attention as a general input representation for computer vision and are often limited to $(x,t)$ or $(y,t)$ cuts.
Our position is that the slicing of the video volume is a design variable: frames are one useful cut among many, STIs are cheap to produce, they turn hard temporal problems into simpler image problems, and they allow existing datasets and frame-based architectures to be reused.
To support research in this direction, we define spacetime images formally, together with their parameters and setup, and we explain what they represent and why the choice of the set of voxels, which we call a \textit{cut}, matters. 
We then illustrate them on real videos and tasks and discuss their advantages and limitations relative to frame-based methods.

\paragraph{Contributions.}

We summarize our contributions as follows: (i) We define STIs and cuts formally, with their parameters and setups, and we show that a family of cuts tiling the video is in bijection with its voxels, so that any dense annotation transfers for free. (ii) We analyze how the choice of the cut shapes an STI, and catalog how pedestrians, deposited objects, camera motion, and dynamic scenes appear in the image and in the frequency domain. (iii) We discuss when STIs help and propose two pilot experiments: we train a segmentation network entirely in the spacetime domain, on transferred annotations, reaching an overall $\Fmeas$ of $0.937$ on the CDnet~2014 background subtraction benchmark without any explicit temporal model, and we recover the pan speed of a camera as the cut whose spectrum is the most concentrated.

\section{Related Work}

\paragraph{Vision-based motion analysis.} 
The analysis of motion can be traced back to the late nineteenth century and the pioneering works of Eadweard Muybridge who captured consecutive phases of animal movement with batteries of still cameras~\cite{Muybridge1887Animal}, and to \'Etienne-Jules Marey, whose chronophotography recorded the successive phases of a movement on a single plate rather than a series of separate pictures~\cite{Marey1894LeMouvement, Braun1992Picturing}.
Motion analysis has since evolved in parallel with technological advances and with the demand for faster and more precise ways of capturing movement in a wide range of settings~\cite{Colyer2018AReview}, and computer vision has inherited from this history the frame as its elementary unit. %
In this position paper, we propose a different way of reading a video that has already been captured. Marey's intuition that one image can hold a whole movement is, in our formalism, just a choice of a cut through the video volume.

\paragraph{Spacetime slices in computer vision.}

Reading a video volume along cuts apart from frames is not an entirely new idea.
Epipolar-plane image analysis stacks the images of a camera moving along a straight line and analyzes slices of the resulting volume, in which the depth of a scene point is read from the slope of its trace~\cite{Bolles1987Epipolarplane}.
In the study of visual motion, a velocity corresponds to an orientation in the $(x,y,t)$ volume, which is the basis of spatiotemporal energy models~\cite{Adelson1985Spatiotemporal} and, later, of the interpretation of such slices in the frequency domain.
On this basis, $(x,t)$ and $(y,t)$ slices have been used to recognize the braided pattern left by a walking person~\cite{Niyogi1994Analyzing, Barnich2010AVirtual}, to track the apparent contours of moving people in real time~\cite{Ricquebourg2000RealTime}, in video-surveillance to detect intrusion~\cite{Zhai2009Virtual} or count people~\cite{Ma2013Crossing}, and to characterize and segment motion through a structure tensor computed on spatiotemporal slices~\cite{Ngo2003Motion}.
More recently, the three orthogonal planes of the volume have been used as complementary views inside otherwise frame-based pipelines, as descriptors for dynamic textures~\cite{Zhao2007Dynamic} or as views sharing 2D convolution kernels for action recognition~\cite{Li2019Collaborative}.
In these works, the slice is extracted from a fixed family of axis-aligned cuts and used alongside the frames.
We instead study the cut itself; we define cuts and their evolution directions in general, including oblique and moving cuts.%

\paragraph{Line-scan sensors and single-line imaging.} 
Line-scan sensors acquire one image line at a time and form a two-dimensional image by stacking successive line measurements, which makes them suitable for high-speed, high-resolution scanning along their scan axis.
They are widely used in industrial inspection on conveyor belts~\cite{Kim2024Design} and in satellite imaging~\cite{Gupta1997Linear}, 
as well as for unwrapping cylindrical surfaces~\cite{Zographos1998Line} and for panoramic imaging and the related 3D reconstruction tasks~\cite{Huang2008Panoramic}.
They also serve for measurements such as defect detection and classification~\cite{Tang2023Review}, position and speed estimation~\cite{Zhao2011Balls}, or the monitoring of calibration patterns to recover vibrations and structural displacements \cite{Nayyerloo2011Seismic}, and the choice of the line to be observed is itself a question in such systems~\cite{Forgacs2025TheDetermination}.
The same construction appears outside industrial vision: kymographs summarize the transport of intracellular cargoes along a fixed path over time, the velocity of a cargo being read from the slope of its trace~\cite{Mangeol2016KymographClear}, and space-time image velocimetry measures the surface velocity of a river from the orientation of the texture produced along a line parallel to the flow~\cite{Fujita2007Development}.
These images are spacetime images obtained for a single cut fixed in advance from known geometry and direction of motion, and analyzed for one scalar measurement.
Our contribution is complementary: we extract families of cuts from ordinary videos, catalog what their content looks like when neither the geometry nor the motion is controlled, and discuss the setups in which a good cut can be predetermined and those in which it cannot.

\section{Spacetime Images (STIs)}
\label{sec:sti}

\subsection{Definition of spacetime images}

Let $\domain = \{0,\dots,W-1\}\times\{0,\dots,H-1\}$ be the spatial domain of a video of width $W$ and height $H$, and let $\timeset = \{0,\dots,T-1\}$ be its temporal domain. 
A video is the map
\begin{equation}
    \video : \domain \times \timeset \rightarrow \mathbb{R}^{C} ,\qquad (\pos, t) \mapsto \video(\pos, t)\in \mathbb{R}^{C} ,
    \label{eq:video}
\end{equation}
where $\pos = (x,y)$ is a pixel spatial position, $t$ a frame index, and $C$ the number of channels. 
The elements of $\domain\times\timeset$ are the \textit{voxels} $\textbf{v}=(\pos, t)=(x,y,t)$ of the video, and a \textit{frame} $\video(\cdot,\cdot,t)$ is a slice of voxels orthogonal to the time axis.

A spacetime image $\sti$ is built by choosing a set of voxels, which we call a \emph{cut}, and by sampling these voxels along an evolution direction. Formally, a cut is an ordered set of $N$ distinct voxels $\{\mathbf{v}_{0},\dots,\mathbf{v}_{N-1}\} \subset \domain\times\timeset$, the evolution direction is a vector $\mathbf{e} \in \mathbb{R}^{3}$ with a non-zero temporal component, and the \emph{spacetime image} (STI) is
\begin{equation}
    \sti(n,k) = \video\big(\mathbf{v}_{n} + k\,\mathbf{e}\big) ,\qquad n \in \{0,\dots,N-1\},\; k \in \{0,\dots,K-1\}.
    \label{eq:stigeneral}
\end{equation}
The first axis of the STI, $n$, indexes the selected voxels and the second axis, $k$, their evolution, the number of evolution steps $K$ being chosen arbitrarily. 
The positions $\mathbf{v}_{n} + k\,\mathbf{e}$ do not need to fall on the voxel grid, in which case $\video$ is evaluated by interpolation, and the cut, $\mathbf{e}$ and $K$ are chosen so that every sampled position remains observable in $\domain\times\timeset$.
Hence, following this general formulation, the cut does not need to be straight nor continuous; it can follow any angle in the $(x,y)$ plane, a curve in the $(x,y,t)$ volume, or be a random set of voxels. 
These cut choices determine which motion patterns the STI will highlight or cancel.
Let us note that the requirement that $\mathbf{e}$ has a non-zero temporal component is what differentiates STIs from the conventional frames, which a purely spatial evolution at a fixed $t$ would reproduce.

A particular family of cuts are the ones where the set of voxels are given by an injective map
$\cut:\{0,\dots,N-1\}\rightarrow\domain$ and the evolution follows the time axis only, $\mathbf{e} = (0,0,1)$, so that
\cref{eq:stigeneral} reduces to
\begin{equation}
    \sti_{\cut} : \{0,\dots,N-1\}\times\timeset \rightarrow \mathbb{R}^{C} ,\qquad
    \sti_{\cut}(n,t) = \video\big(\cut(n), t\big) .
    \label{eq:sti}
\end{equation}
Among this family, the \emph{vertical} cuts at position $x$, $\cut^{\mathrm{V}}_{x}(n) = (x,n)$, with $n\in\{0,...,H-1\}$, one per column, and \emph{horizontal} cuts at position $y$, $\cut^{\mathrm{H}}_{y}(n) = (n,y)$, with $n\in\{0,...,W-1\}$, one per row, generate
$(y,t)$ and $(x,t)$ images, as illustrated in \cref{fig:XYT cube rainbow}. 
For an oblique cut, $\cut$ is obtained by rasterizing the corresponding line in $\domain$.
Let us note that this particular case of STIs can also be read directly from a line-scan camera by stacking its successive acquisitions.

\subsection{STI properties}

\paragraph{Tiling property.} 
A family of cuts $\cutset = \{\cut_{1},\dots,\cut_{J}\}$ \emph{tiles} the video when the images of its cuts form a partition of $\domain\times\timeset$. 
The vertical family, with one cut per column ($j\in\{0,...,W-1\}$), and the horizontal family, with one cut per row ($j\in\{0,...,H-1\}$), both tile the video, so in either case
\begin{equation}
    \Phi : \big(j, n, t\big) \mapsto \big(\cut_{j}(n), t\big)
    \label{eq:bijection}
\end{equation}
is a bijection between the voxels of the STIs and those of the video. 

This tiling property leads to three consequences. 
First, a video is rebuilt exactly from the STIs of a tiling family, and only partially from a family whose cuts do not cover $\domain\times\timeset$.
Second, 
any pixel-wise annotation $\mathcal{A}: \domain\times\timeset \rightarrow \mathcal{L}$, with $\mathcal{L}$ a label set, transfers to the spacetime domain at no cost,
\begin{equation}
    \mathcal{A}_{\cut}(n,t) = \mathcal{A}\big(\cut(n), t\big) ,
    \label{eq:annotation}
\end{equation}
and back through $\Phi^{-1}$. 
This matters because a model trained on frames might not tolerate the distortions that a cut introduces and might have to be retrained rather than reused, which the transfer makes possible without any new annotation effort.
Third, a pixel-wise evaluation score computed over the STIs of a tiling family covers exactly the voxels of the corresponding frame-based scores, so both are directly comparable.

\paragraph{Sampling rate and aspect ratio.}
The second axis of the spacetime image can be compressed or dilated by a sampling rate $\tau \in \mathbb{R}^{+}$, giving $\sti^{\tau}_{\cut}(n,k) = \sti_{\cut}(n,\tau k)$, which  amounts to choosing $\mathbf{e} = (0,0,\tau)$ in \cref{eq:stigeneral} and requires interpolation whenever $\tau k$ is not an integer. The imprint of the object will therefore change with $\tau$, which can make it detectable by frame-based detectors. %

\paragraph{Fixed and open setups.} \label{sec:Setups}

When the camera position is \emph{fixed}, or its motion known, and the nature, typical speed and direction of the objects of interest are known, it is possible to predetermine the cut, or set of cuts, \textit{i.e.,} the orientations, positions, and sampling rates, that will optimize or simplify the resolution of the downstream task.
This is the case for car traffic sequences, where the lanes fix the direction of motion, so that a cut perpendicular to the trajectories recovers the shape of every vehicle crossing it. %
In an \emph{open} setup, where the object motion is not known beforehand, as for a hand-held or moving camera, or where there are not predefined trajectories, as in an upward sky view with flying drones, it may then be useful to use various cuts, random or systematically chosen. 
If the speeds of the objects vary a lot, several sampling frequencies are also required. 
A particular case of solving problems in this open setup is to take ``all'' possible cuts into account, for example a set of parallel cuts that cover all the voxels of the video, or enough multiple cuts with various orientations which may intersect and provide redundant information, but also do cover all voxels.

\section{Reading spacetime images}
\label{sec:reading}

This section explains what can be recognized in an STI and why STIs look the way they do on different video sequences. 
All images are produced by \cref{eq:sti}, with the cuts sampled as described in \cref{sec:sti} and without any processing beyond the extraction itself. 
We then analyze such images in the frequency domain, which exposes the characteristics that \cref{sec:experiments} later exploits.

\subsection{Choosing a meaningful cut} \label{sec:Cuts example}

\begin{figure}[ht]
    \centering
    \includegraphics[width=0.9\textwidth]{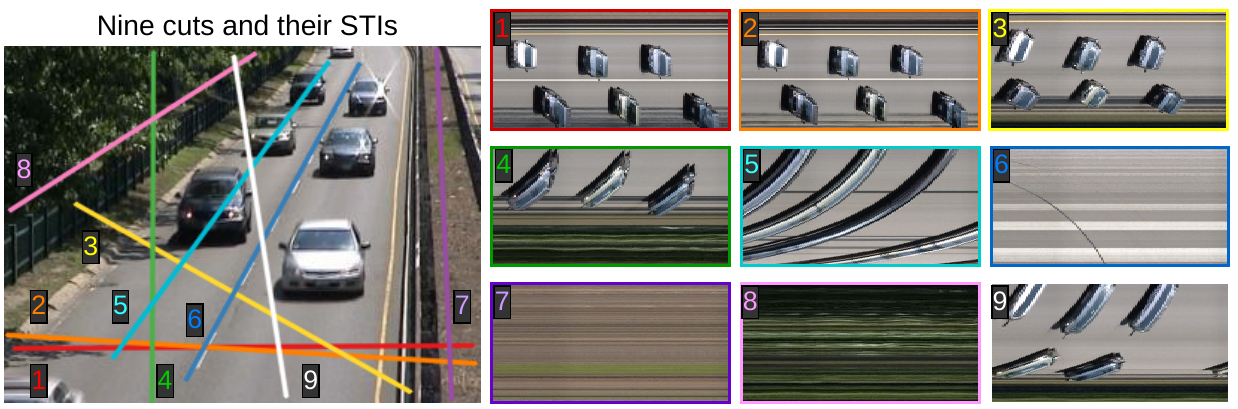}
    \caption{\textbf{Cut design choice.} We showcase nine straight cuts on the same traffic sequence. As can be seen, the direction of the cut heavily influences the content of the STI, including the appearance of certain objects in the scene, more or less distorted.}
    \label{fig:Various cuts}
\end{figure}

Any cut can be used to create an STI, but some are more meaningful than others, depending on the video, and on the objects of interest and their motion. 
\cref{fig:Various cuts} shows the STIs obtained from nine straight cuts on the same sequence captured on a highway. %
Cuts parallel to the image edges are the intuitive choice, but their usefulness depends on the scene: the horizontal cut~1 is crossed by every vehicle, and the resulting STI carries both the shape of the objects and their temporal separation, mapped to a spatial distance, whereas the vertical cut~4 misses the trajectories of the cars in the right lane. 
A cut perpendicular to the trajectory of the objects, when that trajectory is known, is another natural choice (cut~2), as the shape of an object is then recovered up to a shear that depends on the lane it follows. %
A more oblique cut trades one lane against the other and straightens the objects of both~(cut~3).
Cuts nearly parallel to a trajectory still detect a motion but do not preserve enough of the shape for recognition (cuts~5 and~6). 
A cut along a lane marking, such as cut~6, would nevertheless reveal a vehicle changing lanes.
A cut in a static region of the video gives a monotone pattern (cut~7), easily distinguished from a cut in a dynamic background such as trees moving with the wind, producing a more structured pattern (cut~8).
Finally, an arbitrary cut (cut~9) may reveal the moving objects that cross it but distort their proportions.  
The important parameter is therefore the angle between the cut and the expected motion, up to the scaling factor $\tau$.

\subsection{Objects, backgrounds, and camera motion}

\begin{figure}[t]
    \centering
    \includegraphics[width=0.95\textwidth]{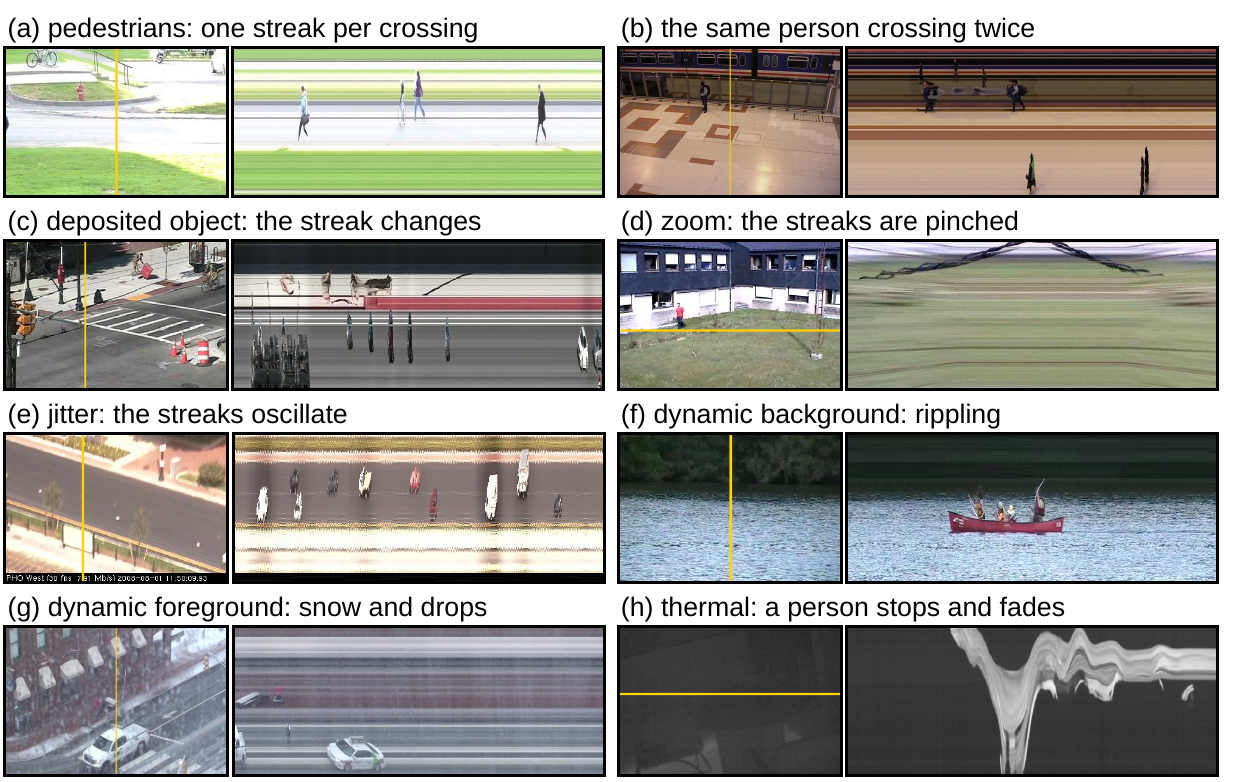}
    \caption{\textbf{A catalog of spacetime images.} For each sequence, the reference frame with the cut \textbf{[left]} and the resulting STI \textbf{[right]}. Moving objects break the streaks, and the way they break them identifies the event.}
    \label{fig:phenomena}
\end{figure}

The simplest STIs are obtained when all objects are rigid and move at similar speeds in similar directions, as in \cref{fig:Various cuts}. \cref{fig:phenomena} considers one cut in each of eight videos with different objects, background and foreground dynamics, and camera motions to illustrate different effects of the cuts.

\paragraph{Pedestrians.}
Unlike a vehicle, a pedestrian changes appearance while walking because the legs alternate and may stop, change direction, or turn back. 
Pedestrians with roughly linear trajectories leave clear imprints~(a), and a person who crosses the cut several times appears several times~(b), so that an STI alone does not distinguish a person coming back from two distinct people.

\paragraph{Changes in the background.}
When an object is deposited, the streak at its position changes color from that instant on~(c), turning an intermittent object motion, notoriously difficult for temporal background models~\cite{Bouwmans2019Deep}, into a discontinuity visible at a glance. 
Conversely, a person who stops moving stops breaking the streaks and merges into the background, as in the thermal sequence~(h), so that an object of interest leaves a background imprint as long as it remains still.

\paragraph{Camera motion.}
Static backgrounds usually appear as parallel lines, and camera motion deforms them in a way that identifies it. 
A zoom compresses and stretches them into pinching patterns~(d).
Jitter makes them oscillate~(e), and since the cut used there is vertical, the oscillation of the streaks measures the vertical component of the motion of the camera, a cut of another orientation being sensitive to another component.
Whereas an individual frame carries almost no information about the motion of the camera, a single STI holds it.

\paragraph{Dynamic backgrounds and foregrounds.}
The canoe of~(f) moves at a constant velocity, and a cut perpendicular to its motion preserves its shape up to a scaling along the time axis, as for the cars of \cref{fig:Various cuts}; the water, however, is a dynamic background whose high frequencies remain visible in the STI. 
Dynamic foregrounds, such as falling snow or drops on a window~(g), appear as lines or curves in front of the objects of interest and complicate their interpretation, similarly to their effects on individual frames.

\subsection{Frequency analysis of spacetime images} \label{sec:Freq analysis}

\begin{figure}[htbp]
    \centering
    \includegraphics[width=0.95\textwidth]{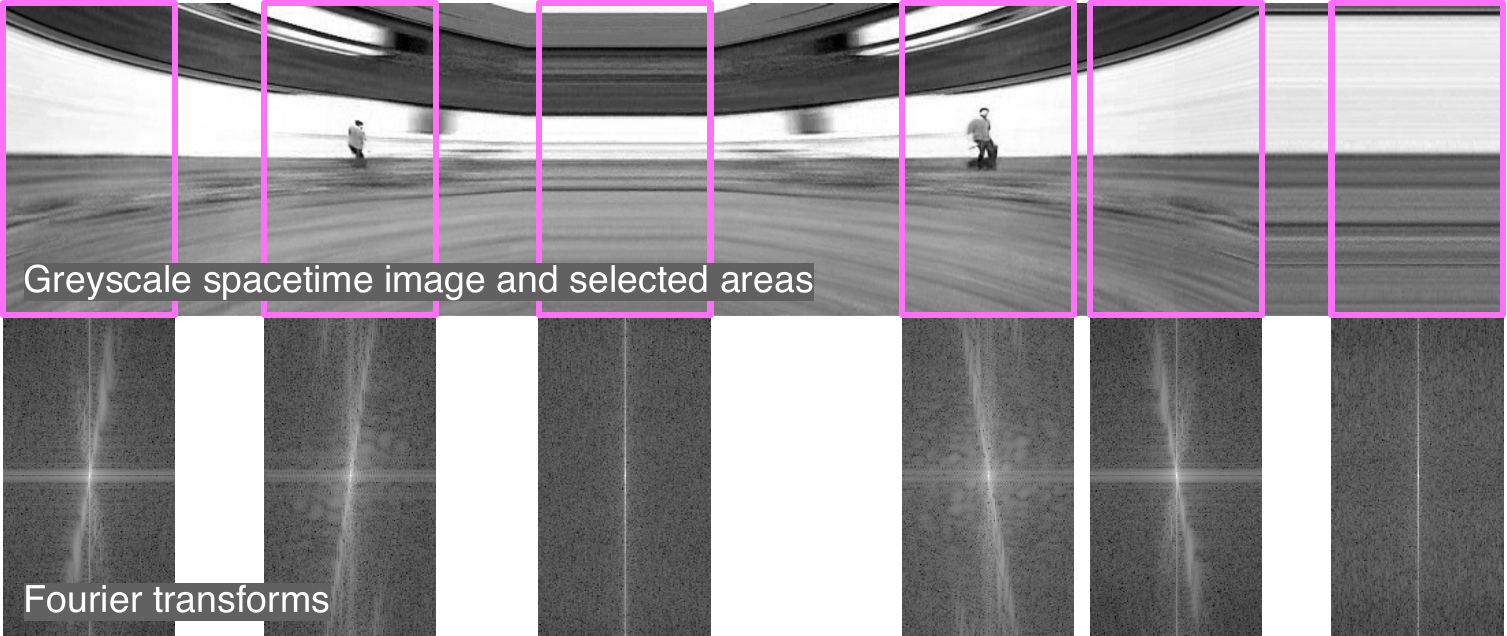}
    \caption{\textbf{Spectral signatures.} Fourier transform computed on windows taken along an STI, extracted from a zoom-in and zoom-out sequence, with log-magnitude spectra. The static zones concentrate their energy on the vertical axis, whereas zooming spreads it over additional orientations and the person adds higher frequency components.}
    \label{fig:Fourier zoom}
\end{figure}

A static scene produces a characteristic pattern in an STI: the voxels of the cut that belong to the background are repeated over time and therefore appear as parallel lines. 
Such a pattern is constant along the time axis, so its energy concentrates on the axis of zero temporal frequency, which is the vertical axis of the spectra of \cref{fig:Fourier zoom}. 
The right end of the STI of \cref{fig:Fourier zoom} corresponds to a static period after the end of the zoom, where both the camera and the building are static, and where little variation beyond noise is present.

In the parts of the image where the camera is zooming, these parallel structures are pinched, diagonal components appear in the STI, and the corresponding spectra show energy in additional orientations. 
The spectra of the zoom-in and zoom-out zones are mirrored regarding each other since the two motions pinch the lines in opposite directions.
A window in which a person crosses the cut likewise shows additional structures and orientations.
The frequency content of an STI therefore reflects changes in the scene, camera motion, and the presence of moving objects, and the concentration of the spectrum on the axis of zero temporal frequency measures how static the content of a cut is, a property that \cref{sec:speed} turns into an estimator.

\section{Experiments}
\label{sec:experiments}

We report two pilot experiments that probe what the change of representation offers: whether a standard vision task can be learned entirely in the spacetime domain (\cref{sec:bgs}), and whether a quantity that frame-based pipelines recover by comparing successive frames, namely the motion of the camera, can instead be read from the choice of a single cut (\cref{sec:speed}).

\subsection{Background subtraction in the spacetime domain}
\label{sec:bgs}

\paragraph{Evaluation setup.}
We use the CDnet~2014 dataset~\cite{Wang2014CDnet}, which contains $53$ videos, grouped in $11$ categories, with pixel-wise foreground ground-truth annotation. 
The annotations are transferred with \cref{eq:annotation} from video voxels to STIs, and pixels outside the spatial or temporal region of interest as well as unknown pixels are ignored, as in the official evaluation. 
Following FgSegNet~\cite{Lim2018Foreground,Lim2020Learning}, we draw $200$ annotated frames per video at random for training and evaluate on all remaining annotated frames.
A frame corresponds to a single column of every STI, so the columns of the non-training frames are masked out of the loss.
This protocol is scene-dependent, since each test scene is also seen during training, which rewards memorization, but it is the protocol under which the supervised methods of the CDnet benchmark report their results, and we follow it for comparability.
We therefore also report a leave-one-video-out experiment, which measures generalization to an unseen video.

\paragraph{Method.}
We train one U-Net~\cite{Ronneberger2015UNet} with an ImageNet-pretrained ResNet-18 encoder~\cite{He2016DeepResidual,Deng2009ImageNet} per cut orientation, on $256\times256$ crops (space $\times$ time) of the vertical and horizontal families, with a masked binary cross-entropy and Dice loss, for $20{,}000$ iterations of batch size $32$. 
At test time, every column and every row is processed with a sliding window, yielding two probability volumes $\hat{Y}^{\mathrm{V}}$ and $\hat{Y}^{\mathrm{H}}$, whose average $\hat{Y} = \frac{1}{2}(\hat{Y}^{\mathrm{V}} + \hat{Y}^{\mathrm{H}})$ we denote \emph{fused}. 
As each family tiles the video, these volumes cover exactly the voxels a frame-based method would produce.

\begin{table}[t]
    \centering
    \caption{\textbf{Background subtraction learned in the spacetime domain} on the $53$ videos of CDnet~2014, under the scene-dependent protocol of~\cite{Lim2018Foreground}: $\Fmeas$ per category for the vertical and horizontal families and their fusion. As in the CDnet evaluation, the overall score is the mean of the per-category scores.}
    \label{tab:bgs}
    \small
    \begin{tabular*}{\linewidth}{@{\extracolsep{\fill}}lccc@{}}
        \toprule
        Category & Vertical cuts & Horizontal cuts & Fused \\
        \midrule
        Baseline & 0.9777 & 0.9697 & 0.9814 \\
        Camera jitter & 0.9591 & 0.9560 & 0.9645 \\
        Bad weather & 0.9502 & 0.9500 & 0.9573 \\
        Dynamic backgr. & 0.9339 & 0.9181 & 0.9437 \\
        Int.\ obj.\ motion & 0.9623 & 0.9422 & 0.9703 \\
        Low framerate & 0.8082 & 0.8206 & 0.8446 \\
        Night videos & 0.8543 & 0.8626 & 0.8788 \\
        PTZ & 0.9309 & 0.9173 & 0.9408 \\
        Shadow & 0.9662 & 0.9622 & 0.9774 \\
        Thermal & 0.9165 & 0.9046 & 0.9276 \\
        Turbulence & 0.8836 & 0.8960 & 0.9197 \\
        \midrule
        \textbf{Overall} & \textbf{0.9221} & \textbf{0.9181} & \textbf{0.9369} \\
        \bottomrule
    \end{tabular*}
\end{table}

\begin{figure}[t]
    \centering
    \includegraphics[width=\textwidth]{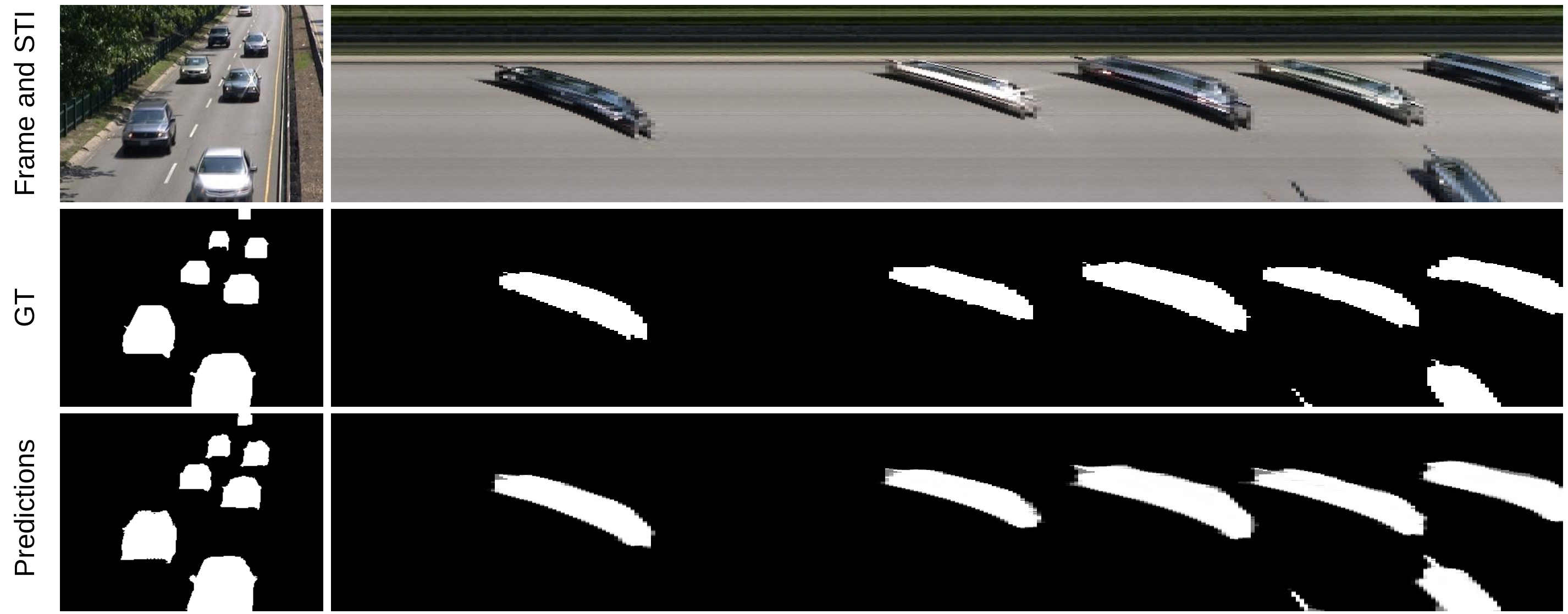}
    \caption{\textbf{Background subtraction obtained in the spacetime domain.} \textbf{[Left]} A test frame, its ground truth, and the foreground segmentations obtained by mapping the spacetime predictions back to the frame domain. \textbf{[Right]} The corresponding STI, its transferred ground truth, and the predicted foreground probabilities.}
    \label{fig:bgs}
\end{figure}

\paragraph{Results.}
\cref{tab:bgs} reports the per-category $\Fmeas$. 
The fused variant reaches an overall $\Fmeas$ of $0.9369$, with a recall of $0.9233$ and a precision of $0.9535$. 
Fusion improves every category over the best single orientation, which is expected since a velocity nearly parallel to one family is nearly normal to the other.
The weakest categories are those where the imprint left on the STI is too thin to be recognized: in the low framerate videos ($0.8446$), objects cross the cut within a few frames and therefore within a few columns, and in the night videos ($0.8788$) the contrast between the object and the background is low.
The network has no explicit temporal module, no recurrence, and no tracker, and yet it segments each imprint as one two-dimensional shape, so that the predictions mapped back to the frame domain are consistent from frame to frame, as illustrated in \cref{fig:bgs}.

\paragraph{Comparison and generalization.}
Under the same protocol, FgSegNet~v2 reports $0.9847$~\cite{Lim2020Learning} with one model per video, whereas we train a single pair for the whole benchmark; ViBe~\cite{Barnich2011ViBe} on these videos gives a mean $\Fmeas$ of $0.62$.
Our purpose is not to compete with per-video models, but to establish that the task is learnable in the spacetime domain, from transferred annotations and without temporal modeling, at a level comparable to supervised frame-based methods.
Scene-dependent protocols also reward memorizing a scene: FgSegNet~v2 falls to $0.3715$ on unseen videos~\cite{Tezcan2020BSUVNet}. 
As a first check that spacetime models do not merely memorize, one leave-one-video-out fold, trained on the $52$ other videos and tested on \texttt{baseline/highway}, gives $\Fmeas = 0.9356$ (vertical), $0.9614$ (horizontal) and $0.9681$ (fused), which is on par with the scene-dependent result on that video and suggests that it learned a transferable representation. 

\subsection{Reading the camera motion from the choice of the cut}
\label{sec:speed}

\paragraph{Principle.}
For a fixed vertical cut, static content produces streaks parallel to the time axis, since the same scene points are sampled at every instant.
If the camera pans, the scene translates in the image, the cut sweeps across it, and the STI becomes a panorama.
The converse gives an estimator: if the cut \emph{moves} with the scene, the same scene points are observed at every instant for the correct motion, and the streaks become parallel to the time axis again. 
In the terms of \cref{eq:stigeneral}, this is a vertical cut whose evolution direction has a spatial component, $\mathbf{e} = (\nu, 0, 1)$, where $\nu$ is a candidate horizontal image velocity in pixels per frame.
Since such a cut leaves the image after a while, it is restarted every $P$ frame, which yields a sawtooth trajectory of the sampled column:
\begin{equation}
    x_{\nu}(t) = x_{c} + \nu\left(\big(t \bmod P\big) - \tfrac{P-1}{2}\right) ,
    \label{eq:sawtooth}
\end{equation}
where $x_{c}$ is the central column, so that $\nu=0$ is the fixed central cut, and the reachable speeds satisfy
$|\nu|(P-1) \le W-1$. 
The estimate is the candidate whose STI is the most static, averaged 
over the periods of the analysis window. 
To do so, we compute the frequency spectrum of the STI images and select the one that has minimal overall non-vertical amplitude.
We sweep $\nu$ with a step of $0.5$~px/frame, refine with a step of $0.05$~px/frame, and interpolate the minimum parabolically.

\begin{figure}[t]
    \centering
    \includegraphics[width=0.95\textwidth]{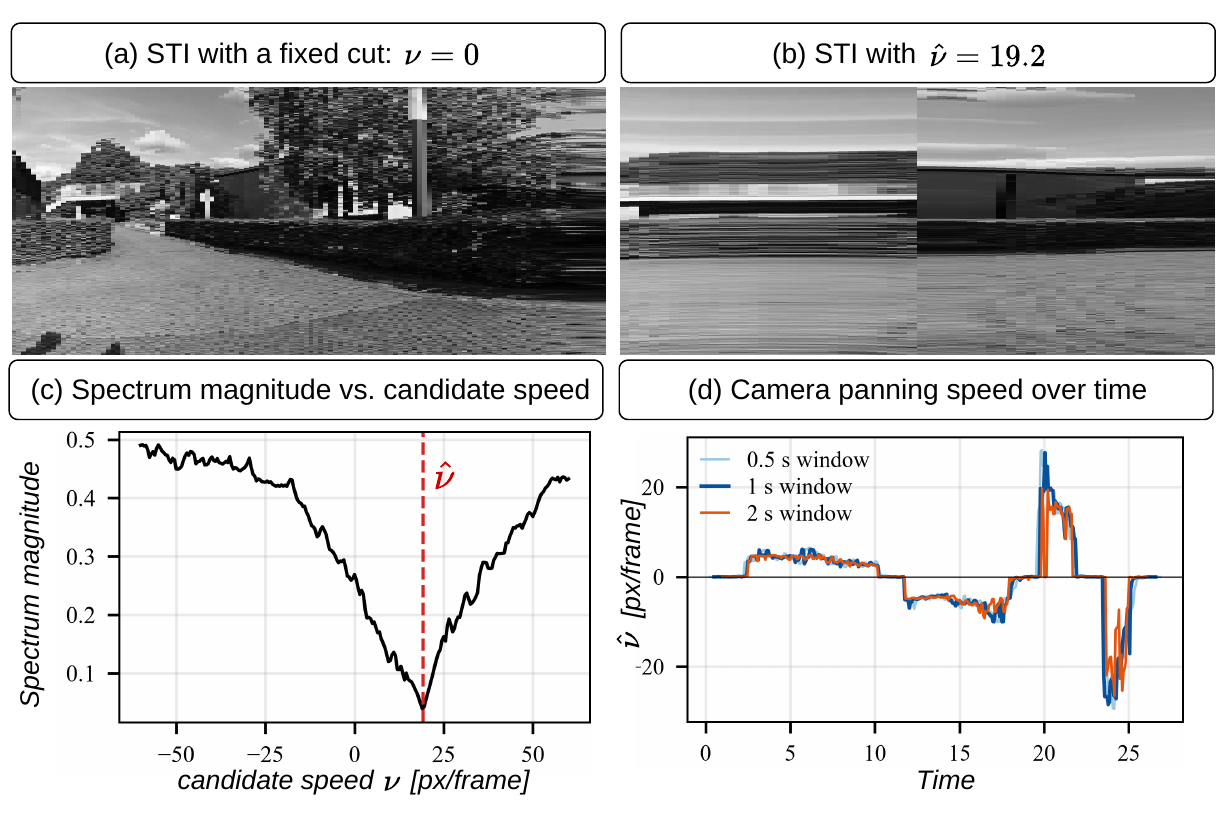}
    \caption{\textbf{Camera speed as the choice of a cut.} \textbf{[Top]} During a fast pan, the STI of a fixed central cut is a panorama of the scene, whereas the sawtooth cut at the estimated speed restores streaks parallel to the time axis. \textbf{[Bottom left]} The spectral amplitude has a single sharp minimum at the image velocity. \textbf{[Bottom right]} Estimated speed along a $27$\,s sequence with pans in both directions at two different speeds.}
    \label{fig:speed}
\end{figure}

\paragraph{Validation on synthetic and real sequences.}
On synthetic pans, built by translating a wide texture assembled from real frames at known sub-pixel speeds, with a window of $1$s and a period $P$ of $0.5$s at $59.94$~fps, the median absolute error is below $0.001$~px/frame for a constant pan at $3.7$~px/frame, a fast pan at $40$~px/frame, and a pan reversing between $-10$ and $+10$~px/frame. 
We also filmed three sequences at $1920\times1080$ and $59.94$~fps with a camera panning on a tripod.
On the sequence of \cref{fig:speed}, the estimate is flat during each slow pan, changes sign as expected, and reaches $\pm 28$~px/frame, \ie about $1700$~px/s; the two others give $-12.3$ to $-4.2$~px/frame and a fast pan peaking at $-34.5$~px/frame. 
Windows of $0.5$ to $1$s follow short pans, whereas longer windows average them out.

\section{Discussion}
\label{sec:discussion}

\subsection{Applications}
\label{sec:applications}

\paragraph{Background modeling.}
In an STI, a static background is a set of streaks parallel to the time axis and a uniformly moving one a set of parallel curves, which makes the separation between background and moving objects a two-dimensional pattern problem, as \cref{sec:bgs} showed on CDnet~2014.
The same structure suits the complementary task of scene background initialization~\cite{Jodoin2017Extensive}, where the background has to be recovered from a sequence that never shows it empty, and the removal of dynamic foregrounds such as rain or snow, which \cref{fig:phenomena}(g) shows as thin lines crossing the streaks.
The spectral signatures of \cref{sec:Freq analysis} also offer a cheap pre-screening since a window whose energy concentrates on the axis of zero temporal frequency contains no moving object and needs no processing.

\paragraph{Detection, counting, and measurement.}
Once the background is removed, each remaining cluster of pixels in an STI localizes one object in space and in time, so that recognition can be applied to the corresponding frame region rather than to the whole video.
Counting follows for objects that cross the cut once ---the ambiguity of \cref{fig:phenomena}(b) being the limiting case.
Object speed can be measured by applying different sampling rates; the sampling rate that straightens the imprint of an object gives its image velocity, which a calibration converts into a physical speed.
Cuts that follow a border turn the monitoring of entries and exits into the same reading, as does cut~6 of \cref{fig:Various cuts} for a vehicle changing lanes.

\paragraph{Camera motion and stabilization.}
\cref{sec:speed} recovers a pan from the choice of the cut; a tilt follows by symmetry with horizontal cuts, whereas a zoom or a rotation requires %
an evolution direction that depends on the voxel, which generalizes \cref{eq:stigeneral}. 
Jitter is the same problem at a shorter time scale: it breaks the streaks into segments, as in \cref{fig:phenomena}(e), and realigning those segments stabilizes the sequence, with one family of cuts per direction of motion to be compensated.

\subsection{Advantages and drawbacks of STIs}
\label{sec:interest}

Three properties make the spacetime domain worth considering.
First, an STI is sensitive to the component of motion normal to the cut and blind to the component parallel to it, so the cut selects a direction of interest instead of filtering it out afterward.
Second, static content collapses into streaks, which removes the scene context that frame-based recognition is known to exploit as a shortcut~\cite{Beery2018Recognition}: a face on a poster is background in an STI, whereas a face detector applied to frames reports it.
Third, a task that only concerns what crosses a given line does not require the whole volume: a single vertical cut samples $H \times T$ voxels out of the $W \times H \times T$ of the video, a fraction $1/W$, so counting objects or estimating their speed analyses a small part of the data, %
whereas a dense task such as background subtraction needs a tiling family and processes every voxel.

The limits are the counterparts of the same properties.
An object that stops moving stops breaking the streaks and merges into the background (\cref{fig:phenomena}(h)), and an object that crosses the cut twice is indistinguishable from two objects (\cref{fig:phenomena}(b)), so an STI alone does not maintain identity.
Motion nearly parallel to the cut is detected but not recognizable, and camera motion deforms every streak, so it has to be estimated and compensated before the background can be treated as static.
Finally, reading an STI requires a temporal window, which makes the analysis offline and introduces a latency of the order of that window.

\section{Conclusion}
\label{sec:conclusion}

This position paper argues that the slicing of the video volume is a design variable.
We defined spacetime images and cuts formally, showed that a family of cuts tiling the video is in bijection with its voxels so that existing dense annotations transfer at no cost, and catalogued how pedestrians, deposited objects, dynamic scenes and camera motion appear in an STI, in the image and in the frequency domain.
Two pilot experiments support the case: a segmentation network trained entirely in the spacetime domain, on transferred annotations and without any explicit temporal module, performs background subtraction on CDnet~2014 close to supervised frame-based methods, and the pan speed of a camera is recovered as the moving cut that makes the static scene static again, identified by a simple criterion on the spectrum of the resulting image.
Spacetime images have long been used in narrow forms, always for one measurement in a controlled setting; what we propose is to treat the cut as a variable of the method rather than as a property of the sensor.
As a result, hard temporal problems become image problems, motion is selected by the geometry of the cut, and existing datasets and 2D architectures can be reused directly.
This opens questions we find worth pursuing: which cuts a method should choose when nothing is known about the scene, how far the estimation of camera motion extends beyond pans, and whether a model can learn the cut rather than receive it.
Frames remain one useful cut among many, and we hope to convince the community that choosing the cut is a decision worth making deliberately.

\subsubsection*{Acknowledgments.}
S. Pi{\'e}rard is funded by grant 2010235 (ARIAC by \href{https://www.digitalwallonia.be/en/}{DIGITALWALLONIA4.AI}) of the SPW EER, Wallonia, Belgium.

\bibliographystyle{splncs04-MVD}
\bibliography{main.bib}

\end{document}